\documentclass[letterpaper, 10 pt, conference]{ieeeconf}  

\IEEEoverridecommandlockouts                              

\usepackage{color}
\usepackage{graphicx,amsmath,amssymb,bm,xcolor}
\usepackage[font=small,skip=0pt]{caption}
\usepackage{subcaption} 
\usepackage[utf8]{inputenc}
\usepackage{breqn} 
\usepackage[noadjust]{cite}
\usepackage[ruled,linesnumbered,norelsize]{algorithm2e}

\usepackage{booktabs}
\usepackage{multirow}
\usepackage{hyperref}
\usepackage{cleveref}
\usepackage{dsfont}
\ifdefined\N                                                                
\renewcommand{\N}{\mathds{N}}                                                
\else
\newcommand{\N}{\mathds{N}}
\fi
\ifdefined\C
\renewcommand{\C}{\mathds{C}}                                             
\else
\newcommand{\C}{\mathds{C}}
\fi

\def\argmin{\mathop{\sf arg\,min}}                                          

\newcommand{\id}{\boldsymbol{I}}                                                

\newcommand{\matb}[1]{ 														
	\begin{bmatrix}
		#1
	\end{bmatrix}
}
\newcommand{\trsp}{{\scriptscriptstyle\top}}									

\newcommand{\nd}[1]{\bm{#1}}
\newcommand{\st}{\mbox{s.t.}}
\newcommand{\norm}[1]{\lVert #1 \rVert}
\newcommand{\bignorm}[1]{\left\lVert #1 \right\rVert}

\Crefname{equation}{Eq.}{Eqs.}
\Crefname{figure}{Fig.}{Figs.}

\title{\LARGE \bf
Efficient and Real-Time Motion Planning for Robotics Using Projection-Based Optimization 
}

\author{Xuemin Chi$^{1*}$, Hakan Girgin$^{2*}$, Tobias L\"ow$^{3}$, Yangyang Xie$^{1}$, Teng Xue$^{3}$, Jihao Huang$^{1}$,
\\ Cheng Hu$^{1}$, Zhitao Liu$^{1\dagger}$, and Sylvain Calinon$^{3\dagger}$
\thanks{$*$ Equal Contribution.}
\thanks{$1$ Authors are associated with the Department of Control Science and Engineering, Zhejiang University.}
\thanks{$2$ Author is with Swiss Cobotics Competence Center, Biel, Switzerland.}
\thanks{$3$ Authors are with the Idiap Research Institute, Martigny, Switzerland
and with EPFL, Lausanne, Switzerland.}
\thanks{$\dagger$ Corresponding authors.}
}

\begin{document}

\maketitle
\thispagestyle{empty}
\pagestyle{empty}

\begin{abstract}
Generating motions for robots interacting with objects of various shapes is a complex challenge, further complicated by the robot’s geometry and multiple desired behaviors. While current robot programming tools (such as inverse kinematics, collision avoidance, and manipulation planning) often treat these problems as constrained optimization, many existing solvers focus on specific problem domains or do not exploit geometric constraints effectively. We propose an efficient first-order method, Augmented Lagrangian Spectral Projected Gradient Descent (ALSPG), which leverages geometric projections via Euclidean projections, Minkowski sums, and basis functions. We show that by using geometric constraints rather than full constraints and gradients, ALSPG significantly improves real-time performance. Compared to second-order methods like iLQR, ALSPG remains competitive in the unconstrained case. We validate our method through toy examples and extensive simulations, and demonstrate its effectiveness on a 7-axis Franka robot, a 6-axis P-Rob robot and a 1:10 scale car in real-world experiments.
Source codes, experimental data and videos are available on the project webpage: 
\url{https://sites.google.com/view/alspg-oc}
\end{abstract}
\section{Introduction}

Many robotics tasks are framed as constrained optimization problems. For example, inverse kinematics (IK) seeks a robot configuration that matches a desired pose while respecting constraints like joint limits or stability. Motion planning and optimal control aim to determine trajectories or control commands that satisfy task-specific dynamics and environmental constraints. Model predictive control (MPC) solves real-time optimal control problems by addressing simplified, short-horizon constrained optimization problems.

Several second-order solvers such as SNOPT \cite{snopt}, SLSQP \cite{slsqp}, LANCELOT \cite{lancelot}, and IPOPT \cite{ipopt}—are commonly used to solve general constrained optimization problems. In robotics, however, most research focuses on solvers tailored to specific problems. For instance, constrained versions of differential dynamic programming (DDP) \cite{constrainedDDP}, iterative linear quadratic regulator (iLQR) \cite{nganga2023second}, TrajOpt \cite{trajopt}, and CHOMP \cite{chomp} are used for motion planning. However, many of these solvers are not open-source, making them difficult to benchmark and improve. Moreover, adapting them for real-time feedback applications, such as closed-loop IK and MPC, often requires significant tuning.

\begin{figure}[t]
    \centering
    \includegraphics[width=\linewidth]{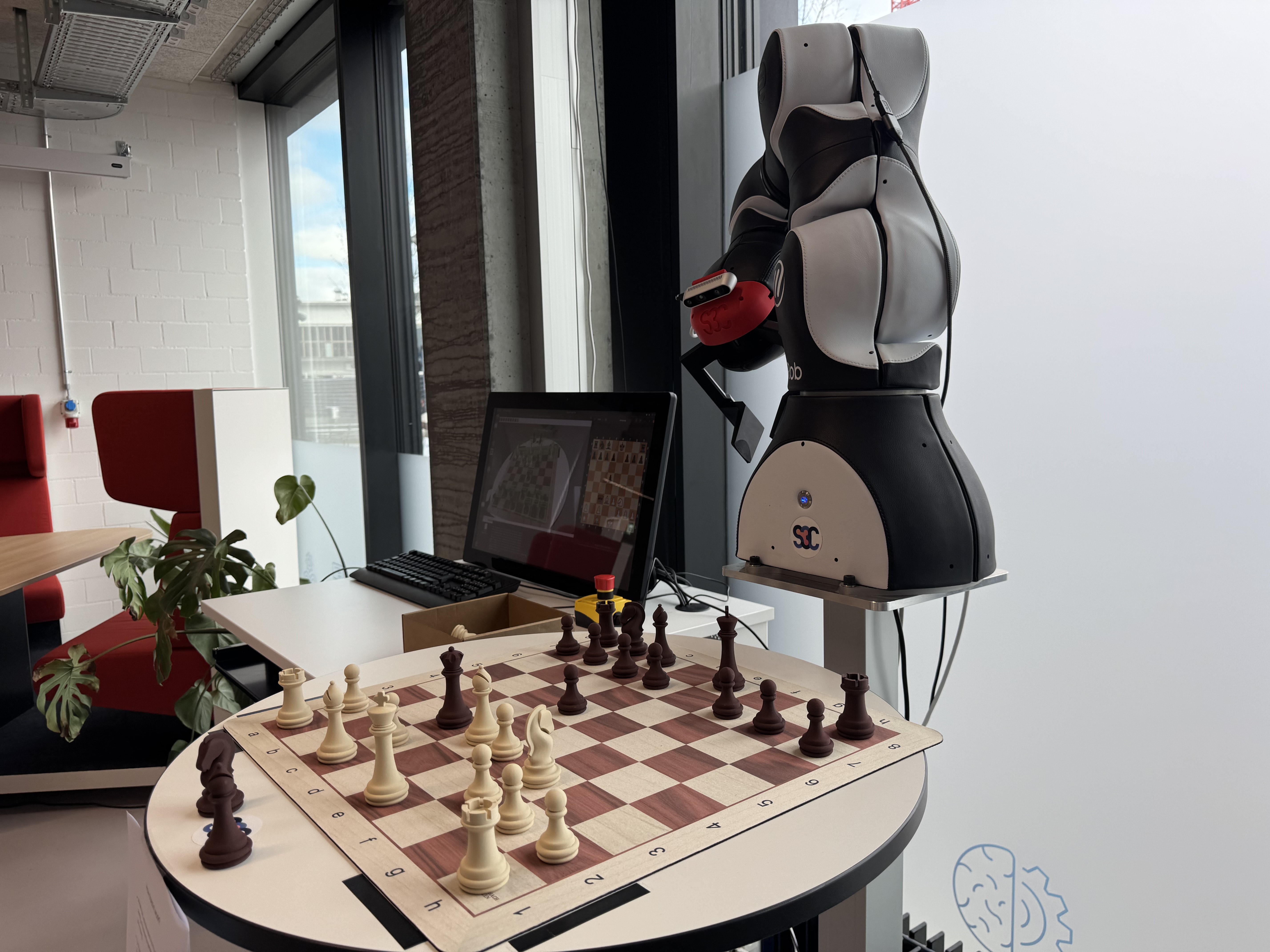}
    \caption{Chess robot setup. SPG-based IK solver validated on a publicly available 5-DOF chess-playing robot, successfully determines correct pick poses autonomously with 100$\%$ accuracy.
    }
    \label{fig:chess_robot}
\end{figure}

We address these challenges by proposing a simple yet powerful solver that can be easily implemented without requiring large memory resources. This solver exploits geometric constraints inherent to many robotic tasks, which can be described using geometric set primitives (see Table \ref{table:projections}). Examples include joint limits, center-of-mass stability, and avoiding or reaching geometric shapes (e.g., spheres or convex polytopes). These constraints can often be framed as projections rather than full constraint formulations. We argue that leveraging these projections, rather than treating constraints generically, can significantly improve solver performance.

One of the simplest algorithms for handling such projections is projected gradient descent, where the gradient is projected to ensure the next iterate remains inside the constraint set. A more advanced version, spectral projected gradient descent (SPG), has demonstrated strong practical performance and is considered a competitive alternative to second-order solvers in various fields \cite{birgin2014spectral}. Extensions of SPG to handle additional constraints using augmented Lagrangian methods have been explored in \cite{andreani2008augmented, birgin2014practical, jia2022augmented}. However, the application of these projection-based methods to popular second-order solvers in robotics has not been widely explored \cite{schmidt2009optimizing}, resulting in a missed opportunity to fully exploit the potential of projections in this domain.

This paper makes the following contributions:

\begin{itemize} \item We propose an efficient projection-based optimization method that leverages geometric constraints in robotics tasks and extend Augmented Lagrangian Spectral Projected Gradient Descent (ALSPG) with a direct shooting method to handle multiple nonlinear constraints.

\item We introduce and integrate various geometric projections including Euclidean projections, polytopic projections, and learning-based projections, into the ALSPG.

\item We validate our approach through numerical simulations on planar arm systems, Franka robot arms, humanoid robots, and autonomous vehicles, providing performance benchmarks and analysis against baseline methods.
We further assess the effectiveness of our method through real-world experiments on 6-axis and 7-axis robotic arms and a 1:10 scale car.
\end{itemize}

\section{Related work}
\label{sec:related}
In optimization, Euclidean projections and the analytical expressions for various projections are fundamental for efficiently solving constrained optimization problems. The general theory for projecting onto a level set of an arbitrary function using KKT conditions is discussed in \cite{projections}. Extensive studies on projection methods and their properties can be found in \cite{bauschke2011convex}, which provides a comprehensive theoretical background. In \cite{usmanova2021fast},
Usmanova \emph{et al.} propose an efficient algorithm for projecting onto arbitrary convex constraint sets, demonstrating that exploiting projections in optimization can significantly improve performance. Further,
Bauschke and Koch \cite{bauschke2015projection} discuss and benchmark algorithms for projecting onto the intersection of convex sets, with Dykstra's alternating projection algorithm \cite{dykstraproject} being a key method for this task.

The simplest algorithm exploiting projections is projected gradient descent, which performs well in many settings.
SPG improves upon this by exploiting curvature information via its spectral stepsizes, leading to faster convergence in many problems. A detailed review of SPG is provided in \cite{birgin2014practical}, highlighting its success in constrained optimization tasks.
In \cite{torrisi2018projected}, Torrisi \emph{et al.} propose to use a projected gradient descent algorithm to solve the subproblems of sequential quadratic programming (SQP). They show that their method can solve MPC of an inverted pendulum faster than SNOPT. Our work is closest to theirs with the differences that we use SPG instead of a vanilla projected gradient descent to solve the subproblems of augmented Lagrangian instead of SQP. Additionally, we propose a direct way of handling multiple projections and inequality constraints, which is not trivial in \cite{torrisi2018projected}. 

In \cite{giftthaler2017projection}, Giftthaler and Buchli propose a projection of the updated direction of the control input onto the nullspace of the linearized constraints in iLQR. This approach can only handle simple equality constraints (for example, velocity-level constraints of second-order systems) and cannot treat position-level constraints for such systems, which is a very common and practical class of constraints in real world applications.

\section{Problem Formulation}
\label{sec:Problem_Formulation}
\begin{table*}[]
\centering
\caption{Projections onto bounded domains, affine hyperplane, quadric and second-order cone.}
\label{table:projections}
\resizebox{0.9\textwidth}{!}
{%
\begin{tabular}{c|c|c|c|c}
\toprule
  & Bounds & Affine hyperplane & Quadratic & Second-order cone \\ \midrule
$\mathcal{C}$ & $l{\leq} x{\leq} u$ & $l{\leq}\nd a^\trsp \nd x {\leq} u$ & $l{\leq}\frac{1}{2}\nd x^\trsp\nd x {\leq} u$ & $\norm{\nd x} {\leq} t$ \\
$\Pi_{\mathcal{C}}$ & $ \begin{cases}
			x & \!\!\text{if } l{\leq} x {\leq} u\\
			u & \!\!\text{if }  x {>} u\\
			l & \!\!\text{if }  x {<} l
		\end{cases}$ & $ \begin{cases}
			\nd x & \!\!\text{if } l{\leq}\nd a^\trsp \nd x {\leq}u\\
			\nd x - \frac{\nd a(\nd a^{\trsp}\nd x - u)}{\norm{\nd a}^2_2} &\!\! \text{if } \nd a^\trsp \nd x {>} u\\
			\nd x - \frac{\nd a(\nd a^\trsp \nd x - l)}{\norm{\nd a}^2_2} &\!\! \text{if } \nd a^\trsp \nd x {<} l
		\end{cases} $ & $\begin{cases}
			\nd x &\!\! \text{if } l{\leq}\frac{1}{2}\nd x^\trsp\nd x {\leq} u\\
			\frac{\nd x\sqrt{2u}}{\norm{\nd x}} &\!\! \text{if } \frac{1}{2}\nd x^\trsp\nd x {>} u\\
			\frac{\nd x\sqrt{2l}}{\norm{\nd x}} &\!\!\text{if } l{>}\frac{1}{2}\nd x^\trsp\nd x
		\end{cases}$ & $\begin{cases}
			(\nd x, t), &\!\! \text{if } \norm{\nd x} {\leq} t\\
			(\nd 0, 0), &\!\! \text{if }  \norm{\nd x} {\leq} {-}t\\
			\frac{\norm{\nd x} + t}{2}(\frac{\nd x}{\norm{\nd x}}, 1) &\!\! \text{otherwise}
		\end{cases} $ \\ \bottomrule
\end{tabular}}
\end{table*}

In this section, we present the definitions of projections used in robotic problems to formulate constraints such as constrained inverse kinematics, obstacle avoidance and other manipulation planning problems.

\subsection{Euclidean Projections}
\begin{figure*}[t]
\centering
\begin{subfigure}{1.5\columnwidth}
\includegraphics[width=\columnwidth]{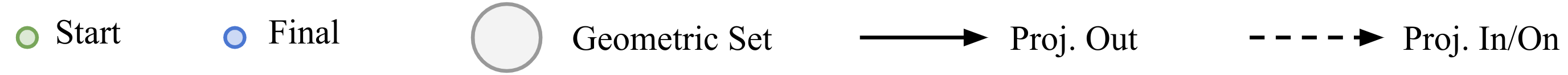}
\end{subfigure}\\
\begin{subfigure}{0.45\columnwidth}
\includegraphics[width=\columnwidth]{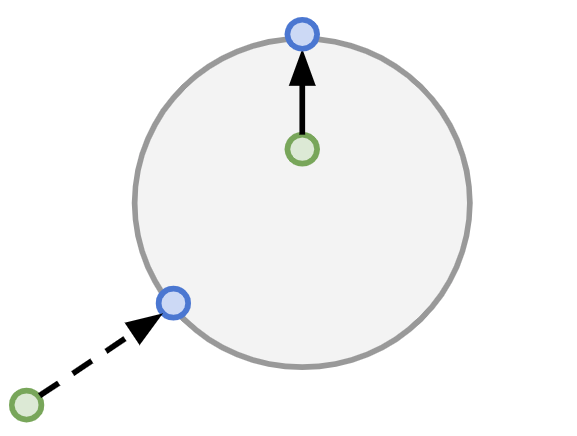}
\caption{}
\end{subfigure}
\begin{subfigure}{0.45\columnwidth}
\centering
\includegraphics[width=\columnwidth]{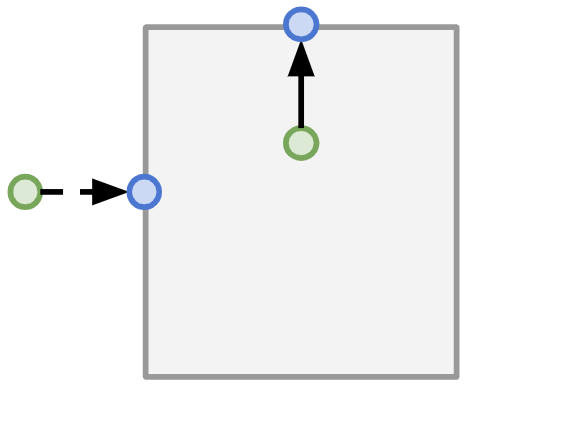}
\caption{}
\end{subfigure}
\begin{subfigure}{0.6\columnwidth}
    \centering
    \includegraphics[width=\columnwidth]{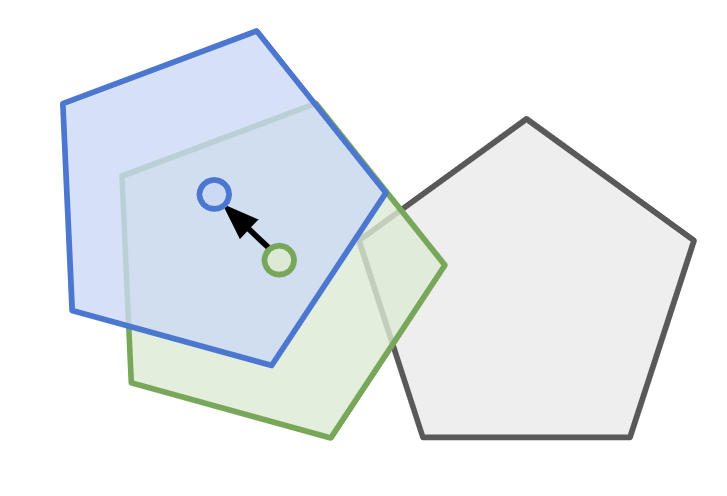}
\caption{}
\end{subfigure}
\begin{subfigure}{0.45\columnwidth}
\centering
\includegraphics[width=\columnwidth]{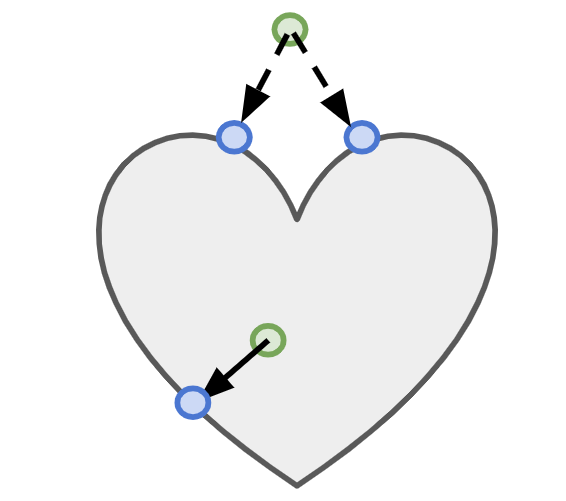}
\caption{}
\end{subfigure}
	\vspace{2mm}
	\caption{Geometric Projections. Projecting outside of a set can be utilized for collision avoidance while projecting inside or onto a set can be employed for goal reaching.
    a) Analytical circle projection.
    b) Analytical box projection. 
    c) Polytopic projection. 
    d) Implicit projection.}
    \label{fig:projection_examples}
\end{figure*}

The solution $\nd x^*$ to the following constrained optimization problem 
\begin{equation}
	\displaystyle \min_{\nd x}  \norm{\nd x - \nd x_0}_2^2 
	\quad\st\quad \nd x \in \mathcal{C}
	\label{eq:Euclidean_projection}
\end{equation}
is called an Euclidean projection of the point $\nd x_0$ onto the set $\mathcal{C}$ and is denoted as $\nd x^*{=}\Pi_{\mathcal{C}}(\nd x_0)$.
This operation determines the point $\nd x \in \mathcal{C}$ that is closest to $\nd x_0$ in Euclidean sense.
For many sets $\mathcal{C}$, $\Pi_{\mathcal{C}}(\cdot)$ admits analytical expressions that are given in \Cref{table:projections}.
Even though, usually, these sets are convex (e.g. bounded domains), some nonconvex sets also admit analytical solution(s) that are easy to compute (e.g. being outside of a sphere).
In cases $\mathcal{C}$ is non-convex, multiple feasible solutions may exist. In such situations, either a strategy for selecting the solution or a convex decomposition method may be required.
Note that many of these sets are frequently used in robotics, from joint/torque limits and avoiding spherical/square obstacles to satisfying virtual fixtures defined in the task space of the robot.  

\subsection{Polytopic Projections}
The geometries of mobile robots often vary, and they can be represented as polytopes defined by hyperplanes.
When considering the robot's geometry, Euclidean projections may not always be applicable. It becomes necessary to account for projections between the robot (modeled as a polytope) and obstacles.
Let the robot's shape be convex, with its occupied space denoted as the set $\mathcal{C}_{r}$.
The geometric center of the polytopic robot, denoted $\nd p \in \mathbb{R}^n$, serves as the point of interest for control and collision avoidance.
We consider the obstacles or goal regions as convex polytopes in $\mathbb{R}^n$ (with $n=2$ or $3$).
If the shapes are not convex, the convex-hulls or convex-decomposition can be employed to approximate them as a collection of convex polytopes.
The Minkowski sum $\mathcal{M}$ between robot and polytopic set $\mathcal{C}$ is defined as:
\begin{equation}
    \mathcal{M} = \mathcal{C}_r \oplus \mathcal{C} = \left\{\boldsymbol{x}=\boldsymbol{x}_1 + \boldsymbol{x}_2 \mid \boldsymbol{x}_1 \in \mathcal{C}_r, \boldsymbol{x}_2 \in \mathcal{C}\right\}
\end{equation}
where $\mathcal{M}$ is the configuration space obstacle for translation movements of robots.
If the geometric center $p\in \mathcal{M}$, the robot and the polytope object intersect.
The problem of projections between two polytopic sets then reduces to projections between the geometric center $\nd p$ and the Minkowski sum $\mathcal{M}$, since $\mathcal{M}$ is composed of hyperplanes, projecting a point out of a polytope, $\nd x^*{=}\Pi_{\mathcal{M}}(\nd x_0)$, can be resolved using the method outlined~\ref{table:projections}, as shown in \Cref{fig:projection_examples}.
The robot's rotational movements can be accounted for by augmenting the Minkowski set with an additional dimension.

\subsection{Implicit Projections}
When the shape of the object is implicit and cannot be expressed using hyperplanes, learning-based techniques can be employed to design the projections. Bernstein polynomial basis functions are efficient for learning the implicit shape. The advantage of this approach lies in the availability of analytical and smooth gradient information.
Assuming the order of the polynomials is $r$ and there are $c$ control points, the matrix form of the implicit shape is $\mathcal{S}:=\boldsymbol{t}^\trsp \boldsymbol{M} \Phi$, where $\boldsymbol{t}\in \mathbb{R}^{r}$ is the time vector, and $\boldsymbol{M} \in \mathbb{R}^{r\times c}$ is the characteristic matrix and $\Phi\in \mathbb{R}^{c}$ represents the control points. 
The analytical gradient $\nabla_{t} \mathcal{S}$ can be computed efficiently.
The implicit projection is demonstrated in \Cref{fig:projection_examples}.

\section{Augmented Lagrangian Spectral Projected Gradient Descent for Robotics}
\label{sec:method}
This section gives the SPG algorithm along with the non-monotone line search procedure. These algorithms are easy to implement without big memory requirements and yet result in powerful solvers.
Next, we give the ALSPG algorithm based on geometric projections. 
Finally, the direct shooting approach is adopted to formulate the constrained optimization problems for robotics.

\subsection{Spectral Projected Gradient Descent}
SPG is an improved version of a vanilla projected gradient descent using spectral stepsizes. Its excellent numerical results even in comparison to second-order methods have been a point of attraction in the optimization literature \cite{birgin2014spectral}. SPG tackles constrained optimization problems in the form of 
\begin{equation}
	\displaystyle \min_{\nd x} f(\nd x)  
	\quad\st\quad \nd x \in \mathcal{C},
\end{equation}
by constructing a local quadratic model of the objective function 
\begin{align*}
f(\nd x)&\approxeq f(\nd x_k) + {\nabla f(\nd x_k)}^\trsp(\nd x - \nd x_k) + \frac{1}{2\gamma_k}\norm{\nd x - \nd x_k}_2^2, \\
&=\frac{1}{2\gamma_k}\norm{\nd x - (\nd x_k-\gamma_k \nabla f(\nd x_k) )}_2^2 + \text{const.},
\end{align*}
and by minimizing it subject to the constraints as
\begin{equation}
	\displaystyle \min_{\nd x}  \frac{1}{2\gamma_k}\norm{\nd x - (\nd x_k-\gamma_k \nabla f(\nd x_k) )}_2^2
	\quad\st\quad \nd x \in \mathcal{C},
\end{equation}
whose solution is an Euclidean projection as described in \eqref{eq:Euclidean_projection} and given by $\Pi_{\mathcal{C}}(\nd x_k-\gamma_k \nabla f(\nd x_k))$. The local search direction $\nd d_k$ for SPG is then given by
\begin{equation}
\nd d_k = \Pi_{\mathcal{C}}(\nd x_k-\gamma_k \nabla f(\nd x_k)) - \nd x_k,
\end{equation}
which is used in a non-monotone line search (\Cref{algo:line_search}) with $\nd x_{k+1}=\nd x_k + \alpha_k \nd d_k$, to find $\alpha_k$ satisfying $f(\nd x_{k+1}) \leq f_{\text{max}} + \alpha_k \gamma_k {\nabla f(\nd x_k)}^\trsp \nd d_k$, where $f_{\text{max}}=\max\{f(\nd x_{k-j} ) \: | \: 0 \leq j \leq \min\{k, M - 1\}\}$.
Non-monotone line search allows for increasing objective values for some iterations $M$ preventing getting stuck at bad local minima. 
A typical value for $M$ is $10$.
When $M=1$, it reduces to a monotone line search.
\begin{algorithm}
	\caption{Non-monotone line search}
	\label{algo:line_search}
	Set $\beta = 10^{-4}$, $\alpha=1$, $M=10$,
	$c = {\nabla f(\nd x_k)}^\trsp \nd d_k$, \\
	$f_{\text{max}}=\max\{f(\nd x_{k-j} ) | 0 \leq j \leq \min\{k, M - 1\}\}$ \\
	\While(){$f(\nd x_k + \alpha\nd d_k) > f_{\text{max}} + \alpha \beta c$}
	{
		
		$\bar{\alpha} = -0.5\alpha^2 c\Big(f(\nd x_k + \alpha\nd d_k) - f(\nd x_k) - \alpha c\Big)^{-1}$ \\
		\eIf {$0.1\leq\bar{\alpha}\leq0.9$}
		{$\alpha=\bar{\alpha}$}{
			{$\alpha=\alpha/2$}}
		
	}
\end{algorithm}

\begin{algorithm}
	\caption{Spectral Projected Gradient Descent}
	\label{algo:spg}
	Initialize $\nd x_k$, $\gamma_k$,
    $\epsilon{=}10^{-5}$, $k{=}0$\;
	\While(){$\norm{\Pi_{\mathcal{C}}(\nd x_k- \nabla f(\nd x_k)) - \nd x_k}_{\infty}>\epsilon$}
	{	
		Find a search direction by
		$\nd d_k = \Pi_{\mathcal{C}}(\nd x_k-\gamma_k \nabla f(\nd x_k)) - \nd x_k$ \\
		Do non-monotone line search using \Cref{algo:line_search} to find $\nd x_{k+1} = \nd x_k + \alpha_{k} \nd d_{k}$ \\
		
		$\nd s_{k+1}=\nd x_{k+1}-\nd x_{k}$ and $\nd y_{k+1}=\nabla f(\nd x_{k+1})-\nabla f(\nd x_k)$\\
		$\gamma^{(1)}=\frac{\nd s_{k+1}^\trsp \nd s_{k+1}}{\nd s_{k+1}^\trsp \nd y_{k+1}}$ and $\gamma^{(2)}=\frac{\nd s_{k+1}^\trsp \nd y_{k+1}}{\nd y_{k+1}^\trsp \nd y_{k+1}}$\\
		
		\eIf{$\gamma^{(1)}<2\gamma^{(2)}$}{$\gamma_{k+1} =\gamma^{(2)}$}
		{{$\gamma_{k+1} =\gamma^{(1)}-\frac{1}{2}\gamma^{(2)}$}}
		
		$k = k+1$ 
	}
\end{algorithm}

The choice of $\gamma_k$ affects the convergence properties significantly since it introduces curvature information to the solver. Note that when choosing $\gamma_k=1$, SPG is equivalent to the widely known projected gradient descent. SPG uses spectral stepsizes obtained by a least-square approximation of the Hessian matrix by $\gamma_k\id$. These spectral stepsizes are computed by proposals
\begin{align}
\gamma_k^{(1)}=\frac{\nd s_k^\trsp \nd s_k}{\nd s_k^\trsp \nd y_k} \quad  \text{and} \quad \gamma_k^{(2)}=\frac{\nd s_k^\trsp \nd y_k}{\nd y_k^\trsp \nd y_k}, 
\end{align}
where $\nd s_k=\nd x_{k}-\nd x_{k-1}$ and $\nd y_k=\nabla f(\nd x_{k})-\nabla f(\nd x_{k-1})$ \cite{birgin2014spectral}.
In the case of the quadratic objective function in the form of $\nd x^\trsp \nd Q \nd x$, these two values correspond to the maximum and minimum eigenvalues of the matrix $\nd Q$. 
The initial spectral stepsize can be computed by setting $\bm{\bar{x}}_{0} = \nd x_{0}-\gamma_{\text{small}}\nabla f(\nd x_0)$ where $\gamma_{\text{small}}$ is $10^{-4}$, and computing  $\bm{\bar{s}}_{0}=\bm{\bar{x}}_{0}-\nd x_{0}$ and $\bm{\bar{y}}_{0}=\nabla f(\bm{\bar{x}}_{0})-\nabla f(\nd x_0)$. Note that this heuristic operation costs one more gradient computation.
The final algorithm is given by \Cref{algo:spg}.

\subsection{Augmented Lagrangian spectral projected gradient descent}
SPG alone is usually not sufficient to solve problems in robotics with complicated nonlinear constraints.
In \cite{jia2022augmented}, Jia \emph{et al.} provides an augmented Lagrangian framework to solve problems with constraints $\nd g(\nd x) \in \mathcal{C}$ and $\nd x \in \mathcal{D}$, where $\nd g(\cdot)$ is a convex function, $\mathcal{C}$ is a convex set, and $\mathcal{D}$ is a closed nonempty set, both equipped with easy projections. 

In this section, we build on the work in \cite{jia2022augmented} with the extension of multiple projections and additional general equality and inequality constraints. The general optimization problem that we are tackling here is 
\begin{equation}
	\displaystyle \min_{\nd x \in \mathcal{D}}  f(\nd x)
	\quad\st\quad \nd g_i(\nd x) \in \mathcal{C}_i, \quad \forall i \in \{1,\hdots ,p\}
	\label{eq:general_problem}
\end{equation}
where $\nd g_i(\cdot)$ are assumed to be arbitrary nonlinear functions.
Note that even though the convergence results in \cite{jia2022augmented} apply to the case when these are convex functions and convex sets, we found in practice that the algorithm is powerful enough to extend to more general cases.

We use the following augmented Lagrangian function
\begin{align*}
\mathcal{L}(\nd x,\{\nd \lambda^{\mathcal{C}_i}, \rho^{\mathcal{C}_i}\}_{i=1}^p ) &= f(\nd x)+ \\
&\hspace{-2cm}\sum_{i=1}^{p}\frac{\rho^{\mathcal{C}_i}}{2}\bignorm{\nd g_i(\nd x) + \frac{\nd \lambda^{\mathcal{C}_i}}{\rho^{\mathcal{C}_i}} - \Pi_{\mathcal{C}_i}\Big(\nd g_i(\nd x) + \frac{\nd \lambda^{\mathcal{C}_i}}{\rho^{\mathcal{C}_i}} \Big)}_2^2
\end{align*}
whose derivative w.r.t. $\nd x$ is given by
\begin{align*}
\nabla \mathcal{L}(\nd x, \{\nd \lambda^{\mathcal{C}_i}, \rho^{\mathcal{C}_i}\}_{i=1}^p) &= \nabla f(\nd x)+ \\
&\hspace{-3cm}\sum_{i=1}^{p}\frac{\rho^{\mathcal{C}_i}}{2}\nabla \nd g_i^\trsp(\nd x)\Big(\nd g_i(\nd x) + \frac{\nd \lambda^{\mathcal{C}_i}}{\rho^{\mathcal{C}_i}} - \Pi_{\mathcal{C}_i}\Big(\nd g_i(\nd x) + \frac{\nd \lambda^{\mathcal{C}_i}}{\rho^{\mathcal{C}_i}} \Big) \Big),
\end{align*}
using the property of convex Euclidean projections derivative $\nabla \norm{\nd g(\nd x)-\Pi(\nd g(\nd x))}_2^2 = \nabla \nd g(\nd x)^\trsp \big(\nd g(\nd x)-\Pi(\nd g(\nd x)\big)$, see \cite{bauschke2011convex} for details.
$\nd \lambda$ is vector of the Lagrangian multipliers and 
$\rho$ is the penalty parameter.
This way, we obtain a formulation that does not need the gradient of the projection function $\Pi_{\mathcal{C}_i}(\cdot)$.
One iteration of ALSPG optimizes the subproblem $\argmin_{\nd x\in\mathcal{D}}\mathcal{L}(\nd x, \{\nd \lambda^{\mathcal{C}_i}, \rho^{\mathcal{C}_i}\}_{i=1}^p)$ given $\{\nd \lambda^{\mathcal{C}_i}, \rho^{\mathcal{C}_i}\}_{i=1}^p$, and then updates these according to the next iterate. Defining the auxiliary function $V(\nd x, \nd \lambda^{\mathcal{C}_i}, \rho^{\mathcal{C}_i}){=}\bignorm{\nd g_i(\nd x)  - \Pi_{\mathcal{C}_i}\Big(\nd g_i(\nd x) + \frac{\nd \lambda^{\mathcal{C}_i}}{\rho^{\mathcal{C}_i}} \Big)}$, the algorithm is summarized in \Cref{algo:ALSPG}. Note that one can define and tune many heuristics around augmented Lagrangian methods with possible extensions to primal-dual methods. 

\begin{algorithm}[t]
	\caption{ALSPG}
	\label{algo:ALSPG}
	Set  $\nd \lambda_{0}^{\mathcal{C}_i}{=}\nd 0$, $\rho_{0}^{\mathcal{C}_i}{=}0.1$, $\epsilon_2 = 10^{-4}$ \\
	\For ()
    {$k = 0 \leftarrow \infty$}
	{
		 $\nd x_{k+1} =\argmin_{\nd x\in\mathcal{D}} \mathcal{L}(\nd x, \{\nd \lambda^{\mathcal{C}_i}, \rho^{\mathcal{C}_i}\}_{i=1}^p)$ with SPG in \Cref{algo:spg} \\
		 
		 \ForEach{$\mathcal{C}_i$}
		 {
		 		$\lambda_{k+1}^{\mathcal{C}_i} = \rho^{\mathcal{C}_i}\Big(\nd g_i(\nd x) + \frac{\nd \lambda^{\mathcal{C}_i}}{\rho^{\mathcal{C}_i}} - \Pi_{\mathcal{C}_i}\Big(\nd g_i(\nd x) + \frac{\nd \lambda^{\mathcal{C}_i}}{\rho^{\mathcal{C}_i}} \Big)\Big) $ \\
		 	
		 	\eIf{$V(\nd x_{k+1}, \nd \lambda_{k+1}^{\mathcal{C}_i}, \rho_k^{\mathcal{C}_i}) \leq V(\nd x_{k}, \nd \lambda_{k}^{\mathcal{C}_i}, \rho_k^{\mathcal{C}_i} ) $}
		 	{
		 		$\rho_{k+1}^{\mathcal{C}_i}{=}\rho_{k}^{\mathcal{C}_i}$ \\
		 	}{
		 		$\rho_{k+1}^{\mathcal{C}_i}{=}10\rho_{k}^{\mathcal{C}_i}$ \\
		 	}
		 
 		}
	\If{$V\left(\boldsymbol{x}_{k+1}, \boldsymbol{\lambda}_{k+1}^{\mathcal{C}_i}, \rho^{\mathcal{C}_i}_{k}\right) < \epsilon_2$}{\text{break}}
	}
\end{algorithm}

\subsection{Optimal Control with ALSPG}
\label{spg:oc}
We consider the following generic constrained optimization problem
\begin{equation}
	\displaystyle \min_{\nd x \in \mathcal{C}_{\nd x}, \nd u \in \mathcal{C}_{\nd u}} c(\nd x, \nd u)
	\quad\st\quad
	\begin{array}{l}
		\nd x = \nd F(\nd x_0, \nd u), \\
		\nd h(\nd x, \nd u) = \nd 0,
	\end{array}
	\label{eq:oc_problem}
\end{equation}
where the state trajectory $\nd x{=}\matb{\nd x_1^\trsp, \nd x_2^\trsp, \hdots, \nd x_t^\trsp, \hdots, \nd x_T^\trsp}^\trsp$, the control trajectory $\nd u=\matb{\nd u_0^\trsp, \nd u_1^\trsp, \hdots, \nd u_t^\trsp, \hdots, \nd u_{T-1}^\trsp}^\trsp$ and the function $\nd F(\cdot,\cdot)$ correspond to the forward rollout of the states using a dynamics model $\nd x_{t+1}{=}\nd f(\nd x_t, \nd u_t)$. We use a direct shooting approach and transform \Cref{eq:oc_problem} into a problem in $\nd u$ only by considering
\begin{equation}
		\displaystyle \min_{\nd u \in \mathcal{C}_{\nd u}} c(\nd F(\nd x_0, \nd u), \nd u)   
		\quad\st\quad
		\begin{array}{l}
		\nd F(\nd x_0, \nd u) \in \mathcal{C}_{\nd x}, \\
		\nd h(\nd F(\nd x_0, \nd u), \nd u) = \nd 0,
	\end{array}
	\label{eq:oc_problem2}
\end{equation}
which is exactly in the form of \Cref{eq:general_problem}, if $\nd g_1(\nd u) = \nd F(\nd x_0, \nd u)$ and $\nd g_2(\nd u) = \nd h(\nd F(\nd x_0, \nd u), \nd u)$. The unconstrained version of this problem can be solved with least-square approaches. However, assuming $\nd x_t\in\mathbb{R}^{m}$, $\nd u_t\in\mathbb{R}^{n}$, this requires the inversion of a matrix of size $Tn\times Tn$, whereas here we only work with the gradients of the objective function and the functions $\nd g_i(\cdot)$. The component that requires a special attention is $\nabla \nd F(\nd x_0, \nd u)$ and in particular, its transpose product with a vector. It turns out that this product can be efficiently computed with a recursive formula (as also described in \cite{torrisi2018projected}), resulting in fast SPG iterations. Denoting $\nd A_t=\nabla_{\nd x_t} \nd f(\nd x_t, \nd u_t)$, $\nd B_t=\nabla_{\nd u_t} \nd f(\nd x_t, \nd u_t)$, and $\nabla_{\nd u} \nd F(\nd x_0, \nd u)^\trsp\nd y = \nd z$ with $\nd y=\matb{\nd y_0, \nd y_1, \hdots, \nd y_t, \hdots, \nd y_{T-1}}$, $\nd z=\matb{\nd z_0, \nd z_1, \hdots, \nd z_t, \hdots, \nd z_{T-1}}$, one can show that the matrix vector product $\nabla_{\nd u} \nd F(\nd x_0, \nd u)^\trsp\nd y$ can be written as 

\begin{align*}
	& \matb{
		\nd B_0^\trsp & \nd B_0^\trsp \nd A_1^\trsp & \nd B_0^\trsp \nd A_1^\trsp \nd A_2^\trsp & \hdots & \nd B_0^\trsp \prod_{t=1}^{T-1}\nd A_t^\trsp\\
		\nd 0 & \nd B_1^\trsp & \nd B_1^\trsp \nd A_2^\trsp & \hdots & \nd B_1^\trsp \prod_{t=2}^{T-1}\nd A_t^\trsp \\
		\vdots & \vdots & \vdots & \ddots & \vdots \\
		\nd 0 & \nd 0 & \nd 0 & \hdots & \nd B_{T-1}^\trsp
}\matb{
\nd y_0 \\ \nd y_1 \\ \vdots \\ \nd y_{T-1}
}\\
&= \matb{
	\nd B_0^\trsp(\nd y_0 + \nd A_1^\trsp \nd y_1 + \nd A_1^\trsp \nd A_2^\trsp \nd y_2 +   \prod_{t=1}^{T-1}\nd A_t^\trsp \nd y_{T-1}) \\
	\nd B_1^\trsp(\nd y_1 + \nd A_2^\trsp \nd y_2 + \nd A_2^\trsp \nd A_3^\trsp \nd y_3 +   \prod_{t=2}^{T-1}\nd A_t^\trsp \nd y_{T-1}) \\
	\vdots \\	
	\nd B_{T-2}^\trsp(\nd y_{T-2} +  \nd A_{T-1}^\trsp \nd y_{T-1})\\
	\nd B_{T-1}^\trsp\nd y_{T-1}
},
\end{align*}
where the terms in parantheses can be computed recursively backward by $\bar{\bm{z}}_{t+1}=(\nd y_{t+1} + \nd A_{t}^\trsp\bar{\bm{z}}_{t})$, $\nd z_t=\nd B_{t-1}^\trsp\bar{\bm{z}}_{t}$ and $\bar{\bm{z}}_{T-1}=\nd y_{T-1}$, without having to construct the big matrix $\nabla_{\nd u} \nd F(\nd x_0, \nd u)^\trsp$.
Note that when there are no constraints on the state and $\nd h(\cdot)=0$, \Cref{eq:oc_problem2} can be solved directly with the SPG algorithm.

\section{Experiments}
\label{sec:experiments}
In this section, we perform extensive simulations and real-world experiments on multiple robotic tasks, including IK problems, motion planning, MPC for a contact-rich pushing task, autonomous navigation and parking tasks.
Real-world evaluations were conducted on a 7-axis Franka robot, a 6-axis P-Rob robot, and a 1:10 scale car.
The motivation behind these experiments is to show that: 1) the proposed way of solving these robotics problems can be faster than the second-order methods such as iLQR and IPOPT; and 2) exploiting projections whenever we can, instead of leaving the constraints for the solver to treat them as generic constraints, increases the performance significantly.

\subsection{Inverse kinematics}
\begin{figure}
	\centering
	\subcaptionbox{\label{fig:ilqr_spg_comparison_setup}}
        [0.48\linewidth]{\includegraphics[width=0.45\columnwidth]{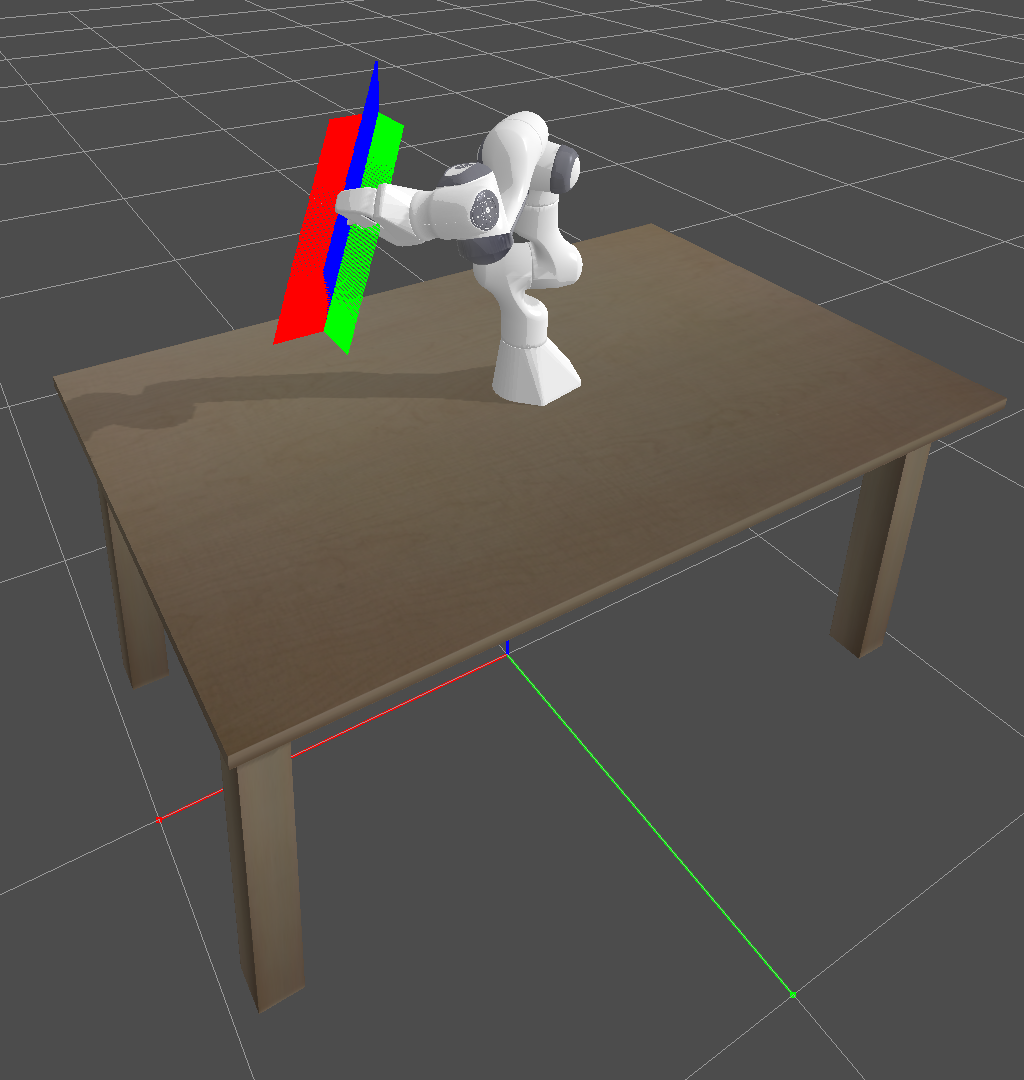}}    
	\subcaptionbox{\label{fig:ilqr_spg_comparison}}
	[0.49\linewidth]{\includegraphics[width=0.49\columnwidth]{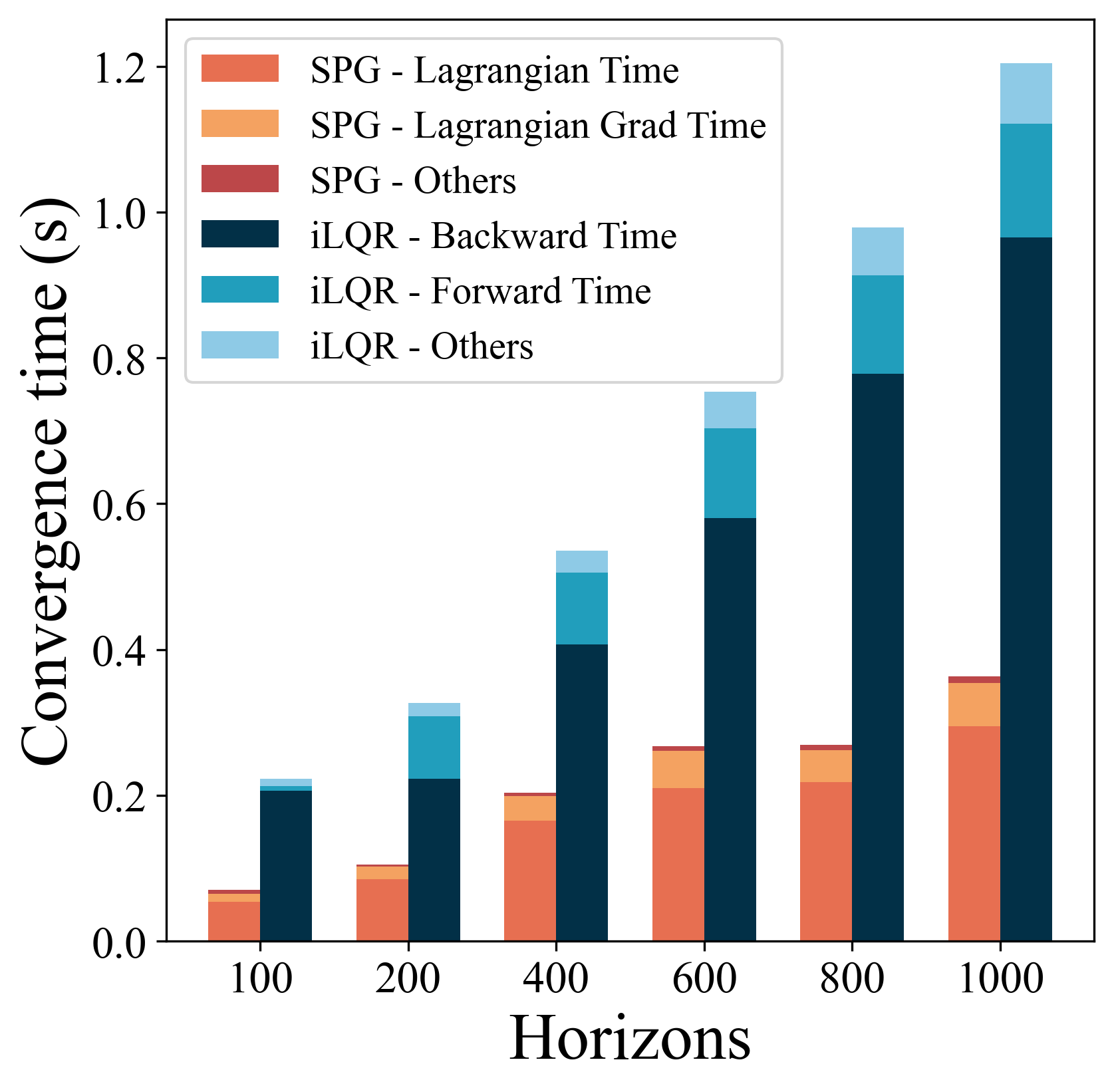}}

\caption{Comparison of iLQR and SPG in terms of convergence time evolution vs the number of horizons.
}
\end{figure}

We first evaluate the performance of SPG compared to iLQR\footnote{The implementation and code of iLQR can refer to \href{https://calinon.ch/codes.htm}{RCFS}.} in a reach planning task using a 7-axis manipulator without constraints.
iLQR is implemented with dynamic programming.
SPG is implemented as detailed in the previous section.
Both implementations are in Python.
The control input is $\ddot{q} \in \mathbb{R}^{7}$ and the states are joint velocities and positions $(\dot{q},q)\in\mathbb{R}^{14}$.
The results on \Cref{fig:ilqr_spg_comparison} show that the convergence time scales linearly with the planning horizons due to the growing number of decision variables.
Notably, SPG scales more efficiently than iLQR, demonstrating its potential for real-time applications.

\begin{table}
\caption{Comparison of constrained inverse kinematics with and without projections.}

\resizebox{0.48\textwidth}{!}
    {%
\begin{tabular}{l|cc}
\toprule
Application         & Num. of fun. eval. & Num of Jac. eval. \\
\midrule
Talos IK without Proj. & $6459.4\pm 3756.8$  &  $3791.79\pm 1061.05$   \\
Talos IK with Proj. & $97.64\pm82.84 $    &   $883.44\pm 81.7$  
\\ 
\bottomrule
\end{tabular}
}
\label{tab:talos_ik}
\end{table}

\begin{figure}[t]
	\centering
	\subcaptionbox{\label{fig:ik_point}}
	[0.25\linewidth]{\includegraphics[width=0.25\columnwidth]{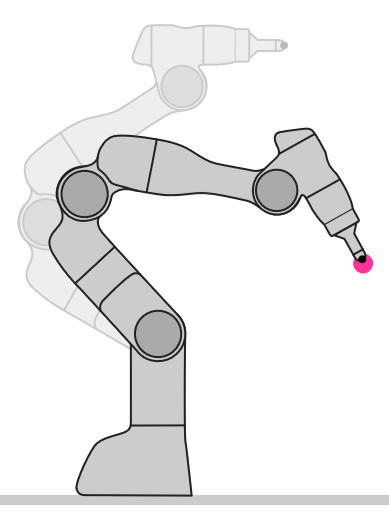}}
	\subcaptionbox{\label{fig:ik_line}}
[0.22\linewidth]{\includegraphics[width=0.22\columnwidth]{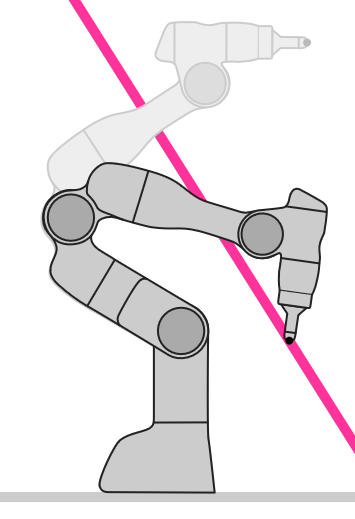}}
	\subcaptionbox{\label{fig:ik_circle}}
[0.252\linewidth]{\includegraphics[width=0.252\columnwidth]{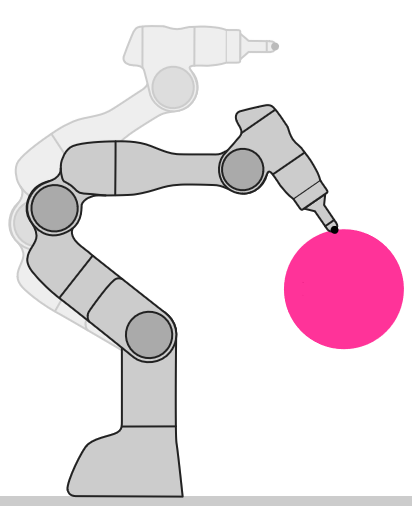}}
	\subcaptionbox{\label{fig:ik_square}}
[0.24\linewidth]{\includegraphics[width=0.24\columnwidth]{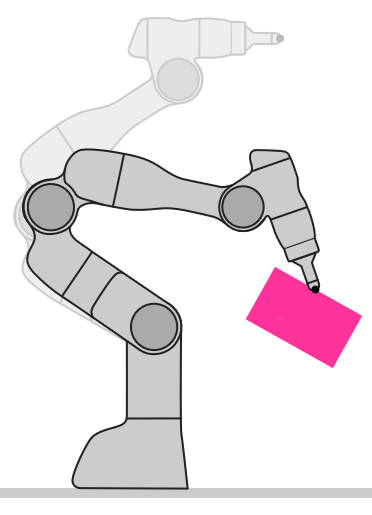}}
\caption{Projection view of inverse kinematics problem. 
(a) Reaching a point (standard IK problem): $\mathcal{C}_{\nd x}=\{\nd x \: | \: \nd x{=}\nd x_d\}$.
(b) Reaching under/above/on a plane (in the halfspace): $\mathcal{C}_{\nd x}=\{\nd x \: | \: \nd a^\trsp \nd x + b{=}0\}$.
(c) Reaching inside/outside/on a circle: $\mathcal{C}_{\nd x}=\{\nd x \: | \: \leq r_i^2 \leq \norm{\nd x-\nd x_d}_2^2 \leq r_o^2\}$.
(d) Reaching inside/outside/on a rectangle: $\mathcal{C}_{\nd x}=\{\nd x \: \: | \: \: \norm{\nd x-\nd x_d}_{\infty, \nd W} \leq L \}$.}
\label{fig:ik_examples}
\end{figure}

A Constrained inverse kinematics problem can be described in many ways using projections. One typical way is to find a feasible $\nd q \in \mathcal{C}_{\nd q}$ that minimizes a cost to be away from a given initial configuration $\nd q_0$ while respecting general constraints $\nd h(\nd q) = \nd 0$ and projection constraints $\nd f(\nd q) \in \mathcal{C}_{\nd x}$
\begin{equation}
	\min_{\nd q \in \mathcal{C}_{\nd q}} \norm{\nd q - \nd q_0}_2^2  
	\quad\st\quad 
	\begin{array}{l}
		\nd h(\nd q) = \nd 0, \\
		\nd f(\nd q) \in \mathcal{C}_{\nd x},
	\end{array}
\end{equation}
where $\nd f(\cdot)$ can represent entities such as the end-effector pose or the center of mass for which the constraints are easier to be expressed as projections onto $\mathcal{C}_{\nd x}$, and $\mathcal{C}_{\nd q}$ can represent the configuration space within the joint limits.
\Cref{fig:ik_examples} shows a 3-axis planar manipulator with $\nd f(\cdot)$ representing the end-effector position and $\mathcal{C}_{\nd x}$ denoting feasible regions.

\begin{figure}
	\centering
	\subcaptionbox{\label{fig:talos}}[0.3\columnwidth]{\includegraphics[width=0.3\columnwidth]{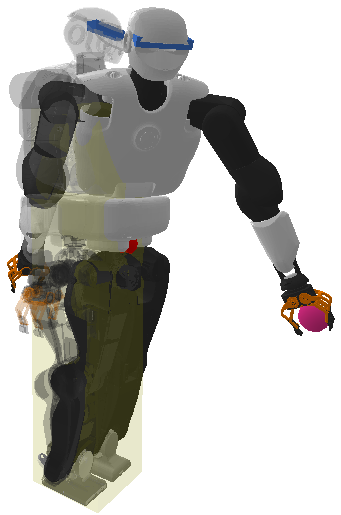}}
	\subcaptionbox{\label{fig:soc_manip}}[0.3\columnwidth]{\includegraphics[width=0.3\columnwidth]{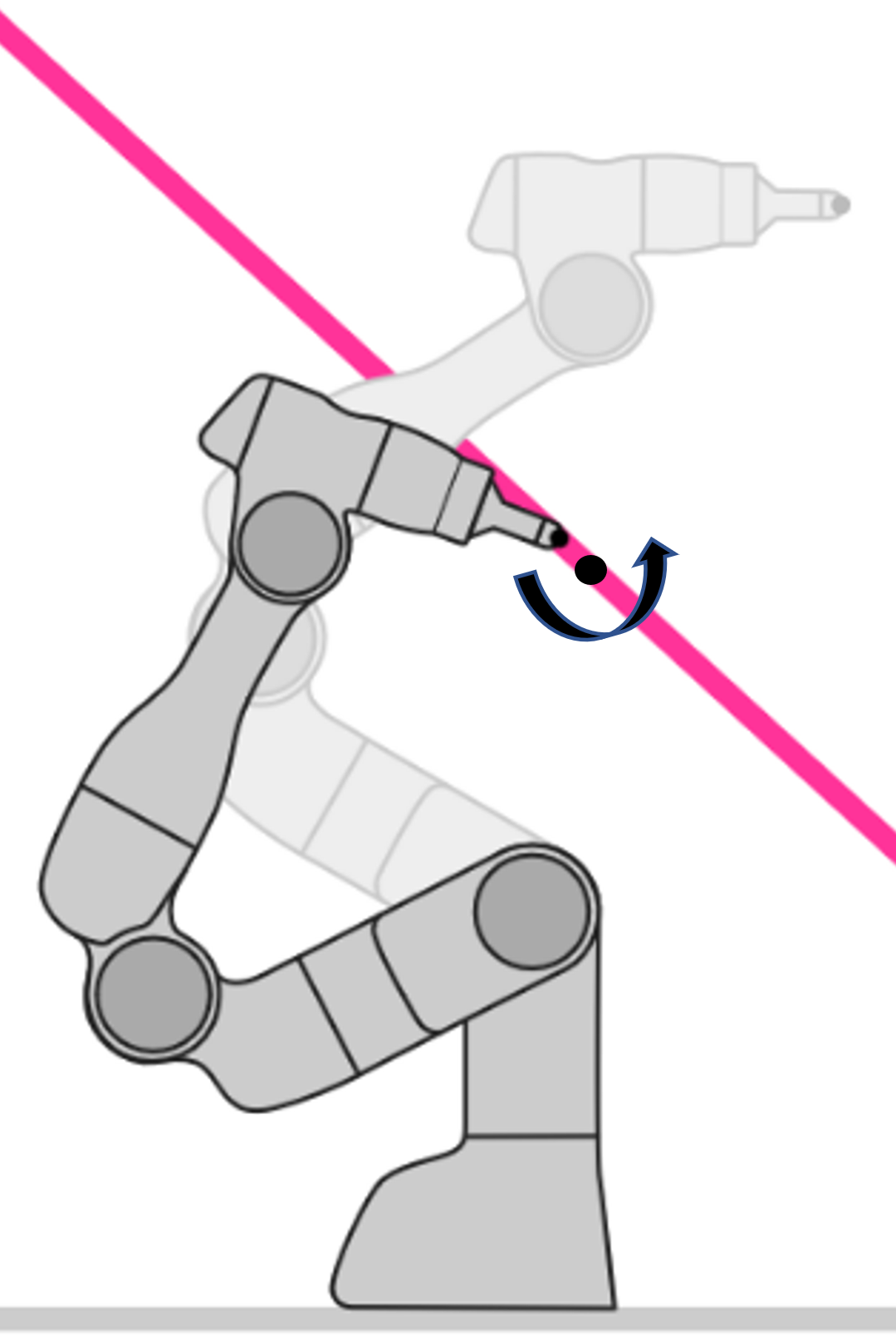}}
	\subcaptionbox{\label{fig:obstacle}}[0.3\columnwidth]{\includegraphics[width=0.3\columnwidth]{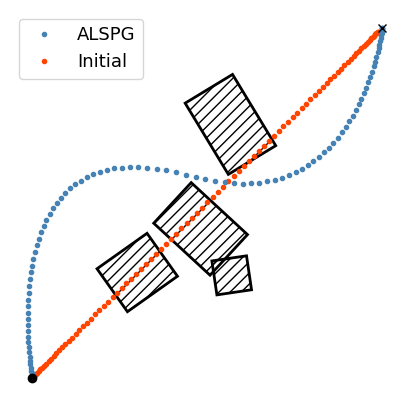}}
	\vspace{2mm}
	\caption{Inverse kinematics and motion planning problems solved with the proposed algorithm. (a) Talos inverse kinematics problem with foot pose, center of mass stability (red point inside yellow rectangular prism) and end-effector inside a (pink) sphere constraints. (b) Robust inverse kinematics solution with $\mathcal{C}_{\nd p}=\{\nd p \: | \:\nd \mu^\trsp \nd p + \Psi^{-1}(\eta) \norm{\nd \Sigma^{\frac{1}{2}}\nd p}_2 \leq 0$\}. (c) Motion planning problem in the presence of 4 scaled and
    rotated rectangular obstacles.
    }
\end{figure}

\textbf{Talos IK}: We tested our algorithm on a high-dimensional (32 DoF) IK problem for the TALOS robot (see \Cref{fig:talos}), subject to the following constraints: i) the center of mass must remain inside a box; ii) the end-effector must lie within a sphere; and iii) the foot position and orientation are fixed. We compared two versions of the ALSPG algorithm: 1) by casting these constraints as projections onto $\mathcal{C}_{\nd x}$; and 2) by embedding all constraints within the function $\nd h(\cdot)$ to assess the direct advantages of exploiting projections in ALSPG.
The algorithm was run from 1000 different random initial configurations for both versions, and we compared the number of function and Jacobian evaluations. The results, shown in Table~\ref{tab:talos_ik}, demonstrate that using projections significantly improves efficiency.

\textbf{Robust IK:} In this experiment, we would like to achieve a task of reaching and staying in the half-space under a plane whose slope is stochastic because, for example, of the uncertainties in the measurements of the vision system. The constraint can be written as $\nd a^\trsp \nd f(\nd q) \leq 0$, where $\nd a \sim \mathcal{N}(\nd \mu, \nd \Sigma)$. 
We can transform it into a chance constraint to provide some safety guarantees probabilistically.
The idea is to find a joint configuration $\nd q$ such that it will stay under a stochastic hyperplane with a probability of $\eta\geq0.5$. This inequality can be written as a second-order cone constraint wrt $\nd f(\nd q)$ as $\nd \mu^\trsp \nd f(\nd q) + \Psi^{-1}(\eta) \norm{\nd \Sigma^{\frac{1}{2}}\nd f(\nd q)}_2 \leq 0$, where $\Psi(\cdot)$ is the cumulative distribution function of zero mean unit variance Gaussian variable. Defining $\nd g(\nd q)=\matb{(\nd \Sigma^{\frac{1}{2}}f(\nd q))^\trsp & \nd \mu^\trsp \nd f(\nd q)}^\trsp$, the optimization problem can then be defined as 
\begin{equation}
	\displaystyle \min_{\nd q \in \mathcal{C}_{\nd q}} \norm{\nd q - \nd q_0}_2^2
	\quad\st\quad 
	\nd g(\nd q) \in \mathcal{C}_{\text{SOC}},
\end{equation}
which can be solved efficiently without using second-order cone (SOC) gradients, by using the proposed algorithm. 
We tested the algorithm on the 3-axis robot shown in \Cref{fig:soc_manip} by optimizing for a joint configuration with a probability of $\eta=0.8$ and then computing continuously the constraint violation for the last 1000 time steps by sampling a line slope from the given distribution. We obtained a constraint violation percentage of around 80\%, as expected.

\subsection{Motion planning and MPC on planar push}
\begin{figure}[t]
	\centering
	\begin{subfigure}{0.48\columnwidth}
		\includegraphics[width=\columnwidth]{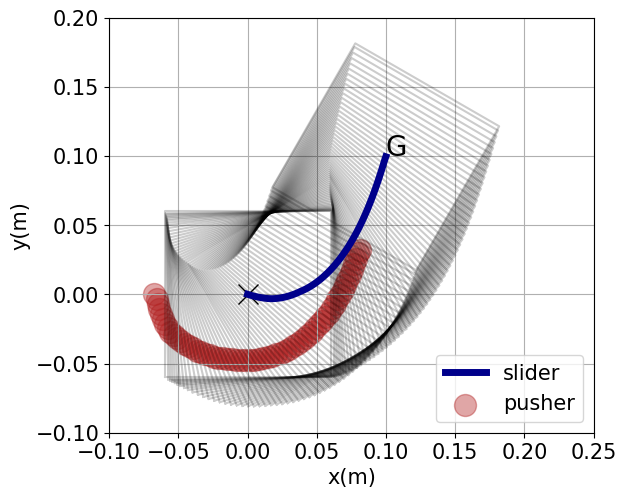}
		\caption{Pusher-slider system path optimized by the proposed algorithm to go from the state $(0,0,0)$ to $(0.1, 0.1, \pi/3)$. Optimal control solved with SPG results in a smooth path for the pusher-slider system.}
	\end{subfigure}\hfill
	\begin{subfigure}{0.48\columnwidth}
		\centering
		\includegraphics[width=0.87\columnwidth]{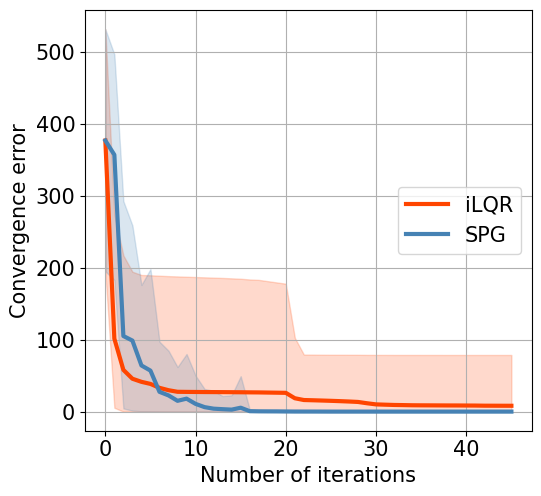}
		\caption{Convergence error mean and variance plot for iLQR and ALSPG motion planning algorithm for 10 different goal conditions starting from the same initial positions and control commands.}
		\label{fig:push_plan_vs}
	\end{subfigure}
	\vspace{2mm}
	\caption{ALSPG algorithm applied to a pusher-slider system.}
	\label{fig:push_spg}
\end{figure}

Non-prehensile manipulation has been widely studied as
a challenging task for model-based planning and control,
with the pusher-slider system as one of the most prominent
examples \cite{Xue23ICRA} (see \Cref{fig:push_spg}).
The reasons include hybrid dynamics with various interaction modes, underactuation and contact uncertainty. In this experiment, we study motion planning and MPC on this planar push system, without any constraints, to compare to a standard iLQR implementation.
The cost function includes the control effort and the $L2$ norm measuring the difference between the final and target configurations. \Cref{fig:push_plan_vs} illustrates the cost convergence for iLQR and ALSPG across 10 different targets for statistical analysis. Although iLQR seems to converge to medium accuracy faster than ALSPG, because of the difficulties in the task dynamics, it seems to get stuck at local minima very easily. On the other hand, ALSPG performs better in terms of variance and local minima. We applied MPC with iLQR and ALSPG with a horizon of 60 timesteps and stopped the MPC as soon as it reached the goal position with a desired precision.
\Cref{table:push_mpc_vs} shows this comparison in terms of convergence time (s), number of function evaluations and number of Jacobian evaluations. According to these findings, ALSPG performs better than a standard iLQR, even when there are no constraints in the problem.

\begin{table}[t]
\caption{Comparison of MPC with iLQR and ALSPG for planar push.}
\resizebox{0.48\textwidth}{!}
{%
\begin{tabular}{l|ccc}
\toprule
Method & Conv. time {[s]} & Num. of fun. eval. & Num. of Jac. eval. \\ 
\midrule
iLQR   &   $14.5 \pm 1.3$             &   $26689.5 \pm 1830.5$               &  $6104.0 \pm 1455.6$              \\
ALSPG  &   $2.9 \pm 0.5$               &      $225.9 \pm 27.4$            &   $78.4 \pm 8.5$                  \\ \bottomrule
\end{tabular}}
\label{table:push_mpc_vs}
\end{table}

\subsection{Motion planning with obstacle avoidance}
\begin{table}[t]
\caption{Comparison of MPC with iLQR and ALSPG for planar push.}
\resizebox{0.48\textwidth}{!}
{%
\begin{tabular}{l|ccc}
\toprule
Method              & Conv. time {[}s{]} & Num. of fun. eval. & Num. of Jac. eval. \\ \midrule
ALSPG with Proj.    &  \textbf{392.0 ± 66.2}         &   \textbf{332.0 ± 60.6}      &    \textbf{187.8 ± 34.6}      \\
SLSQP with Proj.    &  679.0 ± 260.0        &   438.4 ± 200.7     &    389.2 ± 164.3             \\
ALSPG without Proj. &  2780.0 ± 530.9       &   571.8 ± 109.9     &    399.0 ± 93.9       \\ \bottomrule
\end{tabular}
}
\label{table:point_collision}
\end{table}

Obstacle avoidance problems are usually described using geometric constraints.
In autonomous parking tasks, obstacles and cars are usually described as 2D rectangular objects.
In this experiment, we take a 2D double integrator point car reaching a target pose in the presence of rectangular obstacles (see \Cref{fig:obstacle}).
We apply the ALSPG algorithm with and without projections to illustrate the main advantages of having an explicit projection function over direct constraints.
The main difference is without projections, the solvers need to compute the gradient of the constraints, whereas with projections, this is not necessary.
In order to understand the differences between first-order and second-order methods, we also compared ALSPG-Proj to AL-SLSQP with projections (SLSQP-Proj.), which is the same algorithm except the subproblem is solved by a second-order solver SLSQP from Scipy~\cite{virtanen2020scipy}. 
The box obstacles allow analytical projections and distance computation.
 We performed 5 experiments, each with different settings of 4 rectangular obstacles and compared the convergence properties. The results are given in \Cref{table:point_collision}.
 The comparison, with and without projections, reports a clear advantage of using projections instead of plain constraints in the convergence properties.
 Although the convergence time comparison is not necessarily fair for SPG implementations as the SLSQP solver calls C++ functions, the comparison of ALSPG-Proj. and  SLSQP-Proj. shows that ALSPG-Proj. still achieves lower convergence time. 
 Additionally, we compare it against the optimization-based collision avoidance (OBCA) algorithm~\cite{zhang2020optimization} that is based on distance computation and IPOPT. 
 The bicycle model is used $\dot{c}_x=v \cos \theta, \dot{c}_y=v \sin \theta, \dot{\theta}=\frac{v}{L} \tan \delta, \dot{v}=a$.
where $L=2.7$ m is the wheelbase length. 
The system control inputs $\boldsymbol{u}=\left[\delta, a \right]$.
Other parameters remain the same as the baseline.
As shown in Table~\ref{tab:alspg_ipopt}, the results further validate the efficiency of our approach.

\begin{table}[t]
\caption{Comparison between IPOPT and ALSPG on 100 random parking tests.}
\label{tab:alspg_ipopt}
\resizebox{0.48\textwidth}{!}
{%
\begin{tabular}{lllcccc}
\toprule
\multirow{2}{*}{Scenarios}        & \multicolumn{1}{c}{\multirow{2}{*}{Algorithm}} & \multicolumn{1}{c}{\multirow{2}{*}{Solver}} & \multicolumn{4}{c}{Com. time} \\ \cline{4-7} 
                                  & \multicolumn{1}{c}{}                           & \multicolumn{1}{c}{}                        & mean          & std          & min          & max          \\ \midrule
\multirow{3}{*}{Vertical Parking} & OBCA                                           & IPOPT                                       & 655           & 167          & 380          & 1027         \\
                                  & Polytopic Proj.                           & ALSPG                                       & \textbf{222}           & \textbf{130}          & \textbf{52}           & \textbf{885}          \\
                                  & Polytopic Proj.                           & SLSQP                                       & 837           & 221          & 204          & 1319         \\ \cmidrule{2-7} 
\multirow{3}{*}{Parallel Parking} & OBCA                                           & IPOPT                                       & 708           & 199          & 368          & 1346         \\
                                  & Polytopic Proj.                           & ALSPG                                       & \textbf{470}           & \textbf{136}          & \textbf{199}          & \textbf{909}          \\
                                  & Polytopic Proj.                           & SLSQP                                       & 942           & 249          & 227          & 1894         \\ \bottomrule
\end{tabular}
}
\end{table}

\subsection{Autonomous Navigation on a 1:10 Scale Car}
\begin{figure*}[t!]
    \centering
    \begin{subfigure}[t]{\linewidth}
        \centering
        \includegraphics[width=\linewidth]{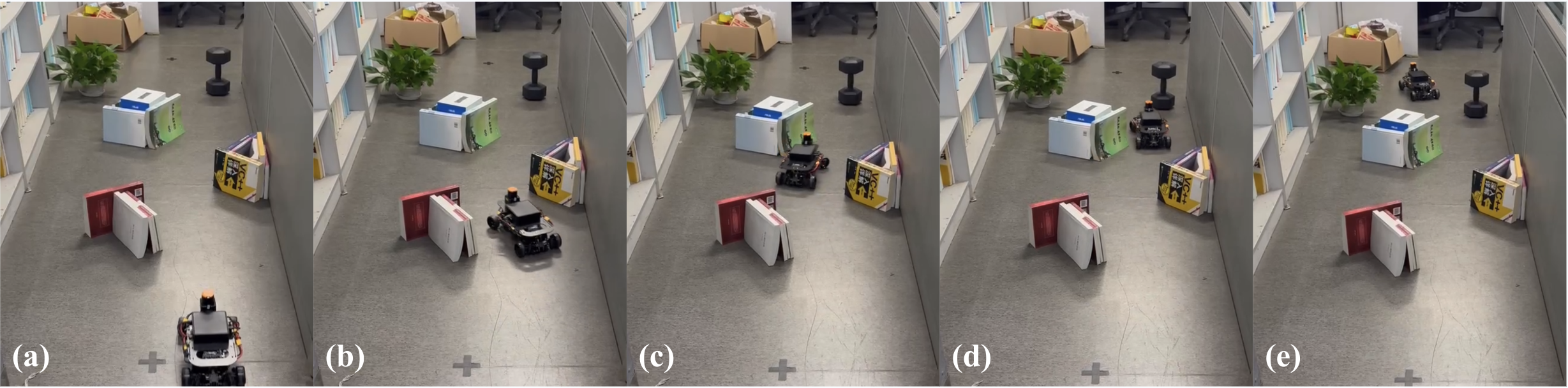}
    \end{subfigure}%

    \caption{ALSPG: Navigation snapshots in an obstacle-cluttered environment.
    a) The car enters the passage.
    b) The car avoids non-convex and triangular obstacles.
    c) A box-shaped obstacle blocks the left side.
    d-e) The car navigates avoiding the box obstacle to reach the goal.}
    \label{fig:navigation_experiments}
\end{figure*}

To further assess the effectiveness of our approach, we tested the ALSPG algorithm on a 1:10 scale vehicle executing a navigation task.
The experiment was conducted on a 1:10 car, using an Intel NUC14 ProU7 as the onboard computer.
The sensor suite includes a Hokuyo UST-10LX LiDAR with a maximum scan frequency of 40 Hz.
Odometry is provided by the VESC. Sensor fusion combines data from the LiDAR, the IMU embedded in the VESC, and odometry, utilizing the Cartographer to localize the vehicle and obtain its state. A pure-pursuit controller is implemented to track the given trajectory.
The projections are constructed through polytopic projections and convex decomposition from section~\ref{sec:method}.
The results shown in Fig.~\ref{fig:navigation_experiments} demonstrate the effectiveness of our method.

\subsection{MPC for Real-Time Tracking on 7-axis Manipulator }
\begin{figure}[t]
	\centering
	\includegraphics[width=\columnwidth]{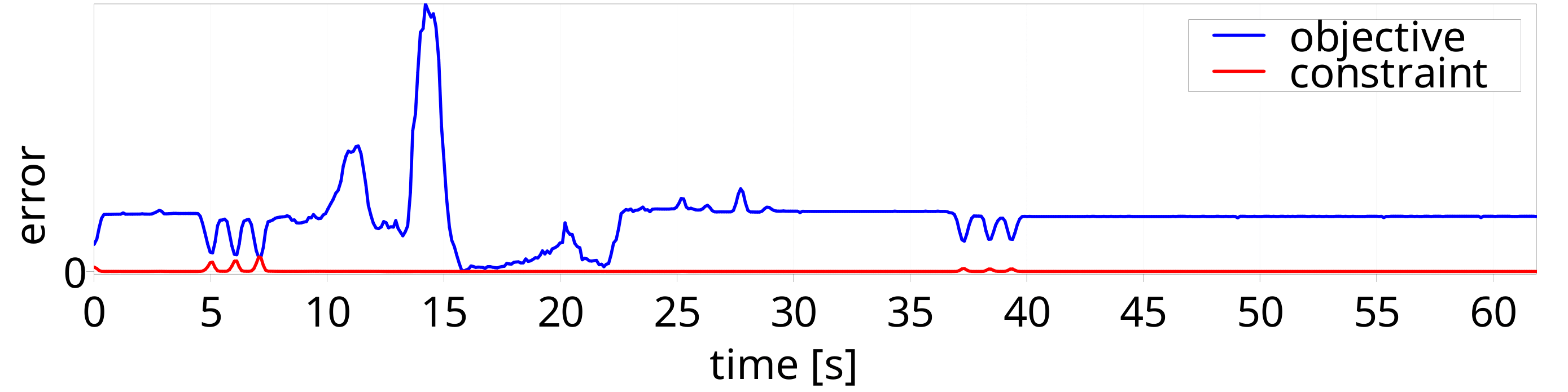}
	\caption{Error in the objective and the squared norm of the box constraint value during 1 min execution of MPC on Franka robot.}
	\label{fig:mpc_box}
\end{figure}

\begin{figure}[t]
	\centering
	\includegraphics[width=0.9\columnwidth]{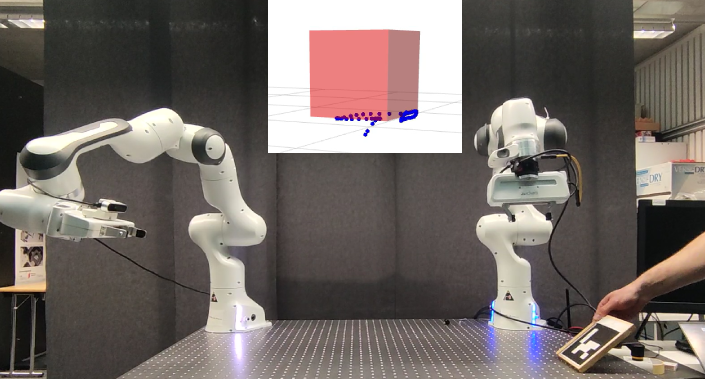}
	\caption{MPC setup for tracking an object subject to box constraints.}
	\label{fig:exp_setup}
\end{figure}

We tested the ALSPG algorithm on the MPC problem of tracking an object with box constraints on the end-effector position of a Franka robot (see \Cref{fig:exp_setup}).
An Aruco marker on the object is tracked by a camera held by another robot.
In this experiment, the goal is to show the real-time applicability of the proposed algorithm for a constrained problem in the presence of disturbances.
In \Cref{fig:mpc_box}, the error of the constraints and the objective function using formulation~\eqref{eq:oc_problem} is given for 1 min. time period of MPC with a short horizon of 50 timesteps.
Between 20s and 30s, the robot is disturbed by the user thanks to the compliant torque controller run on the robot.
We can see that the algorithm drives smoothly the error to zero, see the accompanying video. 

\subsection{Chess Robot}
We validated our algorithm through a three-month experimental study using a 6-axis P-Rob robot from F$\&$P Robotics to solve an inverse kinematics problem (position and orientation) with one degree of freedom (DoF) in orientation left unconstrained. The task involved grasping chess pieces, where the robot autonomously determined its orientation around the z-axis using SPG, compensating for workspace limitations that prevented full 6-DoF positioning, as shown in Fig.~\ref{fig:chess_robot}. The quaternion error, expressed via log-mapping, introduced nonlinearity and complexity, while joint limit projections ensured feasibility. After three months of public demonstrations, the system achieved a $100\%$ success rate.

\section{Conclusion}
\label{sec:spg_conclusion}
In this work, we presented a fast first-order constrained optimization framework based on geometric projections, and applied it to various robotics problems ranging from inverse kinematics to motion planning.
We showed that many of the geometric constraints can be rewritten as a logical combination of geometric primitives onto which the projections admit analytical expressions.
We built an augmented Lagrangian method with spectral projected gradient descent as a subproblem solver for constrained optimization.
We demonstrated: 1) the advantages of using projections when compared to setting up the geometric constraints as plain constraints with gradient information to the solvers; and 2) the advantages of using spectral projected gradient descent based motion planning compared to a standard second-order iLQR and IPOPT algorithm through different robot experiments. 
Sample-based MPC have been increasingly popular in recent years thanks to their fast practical implementations, despite their lack of theoretical guarantees.
In contrast, second-order methods for MPC require a lot of computational power but with somewhat better convergence guarantees. We argue that ALSPG, being already in between these two methodologies in terms of these properties, promises great future work to combine it with sample-based MPC to further increase its advantages on both sides.

\bibliographystyle{IEEEtran}
\bibliography{bib_spg}

\end{document}